\pdfoutput=1
\documentclass[11pt]{article}

\usepackage[]{ACL2023}

\usepackage{times}
\usepackage{latexsym}
\usepackage{amssymb}
\usepackage{graphicx}
\usepackage{cleveref}
\usepackage{placeins}
\crefformat{section}{\S#2#1#3}
\crefformat{subsection}{\S#2#1#3}
\crefformat{subsubsection}{\S#2#1#3}
\crefrangeformat{section}{\S\S#3#1#4 to~#5#2#6}
\crefmultiformat{section}{\S\S#2#1#3}{ and~#2#1#3}{, #2#1#3}{ and~#2#1#3}
\crefmultiformat{subsection}{\S\S#2#1#3}{ and~#2#1#3}{, #2#1#3}{ and~#2#1#3}
\usepackage{refstyle}
\Crefformat{figure}{#2Fig.~#1#3}
\Crefmultiformat{figure}{Figs.~#2#1#3}{ and~#2#1#3}{, #2#1#3}{ and~#2#1#3}
\Crefformat{table}{#2Tab.~#1#3}
\Crefmultiformat{table}{Tabs.~#2#1#3}{ and~#2#1#3}{, #2#1#3}{ and~#2#1#3}
\Crefformat{appendix}{Appx.~\S#2#1#3}
\crefmultiformat{appendix}{Appx.~\S#2#1#3}{ and~#2#1#3}{, #2#1#3}{ and~#2#1#3}
\crefformat{algorithm}{Alg.~#2#1#3}

\usepackage[T1]{fontenc}
\usepackage[utf8]{inputenc}

\usepackage{microtype}

\usepackage{inconsolata}
\title{How well do Large Language Models perform in Arithmetic tasks?}

\author{ Zheng Yuan$^{1}$ \space\space Hongyi Yuan$^{12}$ \space\space Chuanqi Tan$^{1}$ \space\space Wei Wang$^{1}$ \space\space Songfang Huang$^{1}$\\
    $^{1}$Alibaba Group \space\space\space\space $^{2}$Tsinghua University  \\
  {\texttt{\{yuanzheng.yuanzhen,chuanqi.tcq,hebian.ww,songfang.hsf\}@alibaba-inc.com}} \\
  {\texttt{yuanhy20@mails.tsinghua.edu.cn}} \\}

\begin{document}
\maketitle
\begin{abstract}
Large language models have emerged abilities including chain-of-thought to answer math word problems step by step \cite{cot}. 
Solving math word problems not only requires abilities to disassemble problems via chain-of-thought but also needs to calculate arithmetic expressions correctly for each step. 
To the best of our knowledge, there is no work to focus on evaluating the arithmetic ability of large language models.
In this work, we propose an arithmetic dataset \textbf{MATH 401} to test latest large language models including GPT-4, ChatGPT, InstrctGPT, Galactica, and LLaMA with various arithmetic expressions and provide a detailed analysis of the ability of large language models.
% We find xxx.
MATH 401 and evaluation codes are released at \url{https://github.com/GanjinZero/math401-llm}.
\footnote{This project is working in progress.}
\end{abstract}

\section{Introduction}
% Large language models (LLMs) demonstrate emergent abilities 
Emergent abilities show in sufficiently large language models (LLMs) \cite{emergent} like chain-of-thought reasoning (COT) \cite{cot}. 
Chain-of-thought reasoning requires LLMs to solve a question by thinking questions step by step which performs well in school math word problems \cite{cot,0shotcot}. 
Recent LLMs are further fine-tuned with instruction tuning \cite{sanh2021multitask,flan,instructgpt} which demonstrates improved COT ability compared to only self-supervised pre-training.
To solve a math word problem, COT disassembles the problem into simple steps. 
For each step, LLMs have to compute correctly based on arithmetic expressions.
Thus, evaluating the arithmetic ability of LLMs is necessary since it is the upper bound of LLMs' ability for solving math word problems.

To this end, we propose an arithmetic dataset named \textbf{MATH 401}.
Different difficulties are contained in this dataset including addition ($+$), subtraction ($-$), multiplication ($\times$), division ($\div$), exponentiation ($\wedge$), trigonometry functions ($\sin,\cos,\tan$), and logarithm functions ($\log,\ln$) of integers, decimals, and irrational numbers ($\pi, e$).
Long arithmetic expressions with brackets are also included which are common in complex math word problems.
Results in Table~\ref{tab:head} show detailed evaluations on OpenAI's GPTs including GPT-4 \cite{gpt4}, ChatGPT\footnote{\url{https://openai.com/blog/introducing-chatgpt-and-whisper-apis}}, GPT-3.5 \cite{instructgpt} and other open-sourced LLMs.
We find that GPT-4 and ChatGPT outperform other models by a large margin in all kinds of arithmetic abilities. 
InstructGPT \cite{instructgpt} and Galactica \cite{taylor2022galactica} do have some arithmetic abilities.
We analyze factors affecting LLMs' arithmetic ability systematically including tokenization (\Cref{sec:Tokenizer}), pre-training (\Cref{sec:pretrain}), prompts (\Cref{sec:prompt}), interpolation and extrapolation (\Cref{sec:inter}), scaling laws (\Cref{sec:sc}), COT (\Cref{sec:cot}), and ICL (\Cref{sec:icl}).

\begin{table*}[h]
    \centering
    \small
    \begin{tabular}{lc|ccccccc|ccccc|ccc}
    \hline
    Model & Size & E & $+-$ & $\times$ & $\div$ & $\wedge$ & Tri & $\log$ & Dec & Neg & Irr & Big & Long & Easy & Hard & All \\
    \hline
GPT-4 &?&$\checkmark$&\textbf{99}&\textbf{67}&\textbf{100}&\textbf{50}&\textbf{68}&\textbf{76}&\textbf{67}&\textbf{67}&\textbf{100}&\textbf{48}&\textbf{96}&\textbf{100}&\textbf{67}&\textbf{84}\\
ChatGPT&?&$\checkmark$&\underline{97}&\underline{65}&\underline{80}&\textbf{50}&\underline{44}&\underline{56}&\textbf{67}&\textbf{67}&\underline{64}&\underline{40}&\underline{68}&\textbf{100}&\underline{49}&\underline{74}\\
InstructGPT&175B&$\times$&\textit{83}&\textit{59}&\underline{80}&36&8&\textit{16}&\underline{64}&\underline{64}&\textit{36}&\textit{4}&\textit{24}&\textit{92}&\textit{22}&\textit{57}\\
CodeX&175B&$\checkmark$&36&27&8&10&8&0&25&25&12&0&0&40&4&22\\
Galactica&120B&$\checkmark$&69&43&\textit{24}&\textit{44}&\textit{16}&0&\textit{57}&\textit{57}&28&0&\textit{24}&78&12&45\\
LLaMA & 65B&$\checkmark$&44&35&8&22&8&0&41&41&20&0&4&52&5&28\\
OPT & 175B&$\checkmark$&33&35&4&12&0&4&25&25&8&0&0&41&2&22\\
GPT-Neox&20B&$\checkmark$&51&48&4&40&4&0&43&43&20&0&8&66&4&35 \\
GLM&130B&$\checkmark$&39&31&8&22&0&0&29&29&24&0&8&46&5&26\\
BloomZ & 176B&$\times$&23&37&12&30&8&0&43&43&20&0&8&39&6&22\\
Bloom & 176B&$\times$&21&37&12&30&0&0&37&37&16&0&0&37&4&20\\
T0++ & 11B&$\times$&6&3&0&6&8&0&3&3&4&0&0&7&2&4\\
Flan-T5 & 11B&$\times$&1&13&4&0&0&0&11&11&8&0&0&6&2&4\\
% Alibaba-0306 & 13B & \checkmark&11&20&0&30&4&0&23&23&8&0&0&24&2&13\\
\hline
    \end{tabular}
    \caption{Arithmetic ability for LLMs measured by accuracy, we only list models with largest parameter counts. E = Euler, Dec = Decimal, Neg = Negative, Irr = Irrational, Big = Big Numbers, Long = Long Expressions.}
    \label{tab:head}
\end{table*}

One may say that the ability to solve arithmetic tasks is not necessary for a large language model.
LLMs can use the calculator API when they need to decode an answer \cite{Schick2023ToolformerLM}.
Arithmetic ability evaluation can be a gauge for general intelligence since mastering arithmetic serves as a fundamental requirement for performing intricate mathematical tasks including symbolic math reasoning \cite{noorbakhsh2021pretrained,10002218} and automatic theorem proofing \cite{Polu2020GenerativeLM,Wu2022AutoformalizationWL}.
% which could be further viewed as .
% This work is to locate the position of LLMs in general mathematics.

\section{Related Works}
\paragraph{Evaluate Math Ability of LLMs}
To show the math reasoning ability of LLMs, \citet{gpt-j,flan,Thoppilan2022LaMDALM} evaluate their models on various math word problems benchmark
\cite{saxton2019analysing,hendrycks2021measuring,cobbe2021gsm8k,Shi2022LanguageMA}.
For newly released LLM ChatGPT, \citet{Shakarian2023AnIE,Frieder2023MathematicalCO} evaluate its mathematical ability independently.
To notice, our paper evaluates ChatGPT using gpt-3.5-turbo-0301 version and GPT-4 using chat UI on March 16th which may have different performances compared to their reported results and future analysis.

\paragraph{Evaluate Arithmetic Ability of LLMs}
\citet{Nogueira2021InvestigatingTL, Wang2021ExploringGA} evaluate pretrained language models on simple arithmetic expressions including addition ($+$) and subtraction ($-$).  \citet{Muffo2022EvaluatingTL} have further tested the multiplication ($\times$) of language models.
They found tokenization \cite{Nogueira2021InvestigatingTL, Kim2021HaveYS} and token frequency \cite{Razeghi2022ImpactOP} are two important factors for language model arithmetic ability. 
% reveal that tokenization is crucial for arithmetic ability.
Compared to previous work, we focus on evaluating \textbf{Large} LMs (with instruction fine-tuning) on comprehensive arithmetic abilities with different types of operators and numbers.
% Furthermore, some LLMs we evaluate have conducted multi-task instruction fine-tuning which may enhance their arithmetic ability.

\section{Evaluation Settings}

\subsection{Arithmetic Expression Settings}
We construct 401 arithmetic expressions to test large language models which include Euler equation ($e^{i\pi}+1=0$) as group 0 and 25 problems each for group 1$\sim$16.
% 16 different groups (group 1$\sim$16) with 25 problems for each group.
If not otherwise mentioned, used numbers are positive integers.

\begin{itemize}
    \item Euler Equation.
    \item Add \& Subtract of two integers within 10.
    \item Add \& Subtract of two integers within 100.
    \item Add \& Subtract of two integers within 1,000.
    \item Add \& Subtract of two integers within 1,000,000,000,000.
    \item Add \& Subtract of two integers within -10$\sim$10.
    \item Add \& Subtract of two decimal numbers within -100$\sim$100.
    \item Multiply two integers within 100.
    \item Multiply two decimal numbers within 10.
    \item Multiply two integers within 100,000.
    \item Division of two integers within 100.
    \item Exponentiation of with integer base within 10 and integer exponent within 2$\sim$4.
    \item Exponentiation of with a decimal number within 10 as the base and a decimal number within 2$\sim$4 as the exponent.
    \item Add, Subtract \& Multiply with one integer within 10 and a common irrational number (i.e. $e$ or $\pi$).
    \item Long arithmetic expressions with brackets, involved integers are all within 100 and operators contain add, subtract, multiply, and division.
    \item Trigonometry functions including $\sin$, $\cos$, and $\tan$. Inputs can be in the format of degrees and radians ($\pi$ can also appear in the inputs).
    \item Logarithm of integers within 1000 of different bases: $2,e,10$.
\end{itemize}

These groups cover mathematical operators used in elementary mathematics. We consider groups 1,2,3,5,6,7,8,11 as \textbf{Easy} queries and all others as \textbf{Hard} queries.
We calculate the results of all arithmetic expressions using built-in functions of Python and round to four decimal places. Examples of expressions are listed in Appendix~\ref{sec:app:math401}.

\subsection{Metrics}
Since LLMs can decode arbitrary contents (which may contain their step-by-step calculation steps), we first ignore decoded numbers in parentheses and preserve the last number decoded by LLMs.
If the decoded number is a fraction, we will convert it to decimal for evaluation except for group 10 which requires calculating division.
To measure the arithmetic ability of LLMs, we use the following metrics to measure their outputs. 

\paragraph{Accuracy} If the difference between the decoded number and the target number is less than $1e-3$, we consider it a correct prediction. Accuracy is calculated based on correct prediction counts.

\paragraph{Relative error} We denote decoded number is $\hat{y}$ and target is $y$. We calculate relative error by:
\begin{equation}
    RE = \min(10, \frac{\|\hat{y}-y\|}{\max(\|y\|, 1)})
\end{equation}
If LLM does not decode any number, we consider $RE=10$.
We truncate the relative error to 10 to prevent that one big mistake dominate the average relative error.

\paragraph{Non-number ratio} If decoded content does not contain any numbers, we consider it a failure. We calculate the non-number ratio based on it.

\subsection{Evaluation Details}
We test GPT-4 by their official chat UI\footnote{\url{https://chat.openai.com/chat?model=gpt-4}}.
Since GPT-4 has limited request counts, we only query GPT-4 with groups that ChatGPT cannot answer correctly.
We test GPT-3.5 (including davinci (CodeX, InstructGPT) and turbo (ChatGPT) series models) \cite{instructgpt,codex} via OpenAI APIs. 
We also test following open-sourced LLMs including Galactica \cite{taylor2022galactica}, GPT from EleutherAI \cite{gpt-j,gpt-neox-20b}, LLaMA \cite{touvron2023llama}, OPT (with instruction learning) \cite{opt}, Bloom (with instruction learning) \cite{scao2022bloom,bloomz}, T0++ \cite{sanh2021multitask}, GLM \cite{zeng2022glm130b} and Flan-T5 \cite{flan}. We also test the smaller versions of the above models.

We test following prompts: $\emptyset$ (i.e. no prompt), ``Calculate:'', ``\$'', ``\$\$'', and ``\textbackslash begin\{equation\}''. The latest three prompts are inspired by that LLMs may be pretrained with \LaTeX\space sources.
We provide three versions of input formats: math texts ($\pi$), plain texts (pi), \LaTeX\space texts (\textbackslash pi). When we use \LaTeX-related prompts, we provide the model with \LaTeX\space texts. When we use other prompts, we will provide math texts if their tokenizers can encode them. Otherwise, we will provide plain text.
For ChatGPT (gpt-3.5-turbo-0301), we test different system-level prompts as instructions: $\emptyset$ (i.e. no prompt), ``You are an accurate calculator.'', and ``You are an accurate calculator, please calculate provided equation to four decimal places.''.
For GPT-4, we only test prompt ``You are an accurate calculator, please calculate provided equation to four decimal places.''.
% we do not give system messages like ``You are a powerful calculator.''
% also give system messages including ``You are a powerful calculator.''

% Different decoding strategies can have apparent influences on decoded texts. 
% Compared with other natural language generation tasks which can have multiple acceptable results, arithmetic tasks usually have only a single correct answer.
We use default decode settings for OpenAI APIs, and we use greedy decoding for all other LLMs.

\begin{table}[t]
    \small
    \centering
    \begin{tabular}{lc|ccc}
    \hline
    Model & Prompt & Acc $\uparrow$ & RE $\downarrow$ & NNR $\downarrow$ \\
    \hline
gpt-4&Cal*4&\textbf{83.54}&\textbf{0.07}&\textbf{0.00} \\
gpt-3.5-turbo-0301&Cal*&\underline{75.06}&\underline{0.14}&\underline{0.50}\\
text-davinci-003&Cal&\textit{56.61}&0.76&2.99\\
code-davinci-002&Eqa&21.7&2.39&11.47\\
galactica-120b&Eqa&45.14&1.30&3.99\\
galactica-30b&Eqa&45.14&\textit{0.69}&\textit{1.75}\\
llama-65b&Eqa&28.43&1.61&4.74\\
opt-175b&Cal&21.70&3.18&21.70\\
gpt-neox-20b&Eqa&35.41&1.19&4.49\\
glm-130b&\$&25.94&1.27&2.74\\
% gpt-j-6b&Cal&27.18&1.55&8.98\\
bloomz-176b&\$\$&22.44&1.50&4.74\\
bloom-176b&\$&20.20&2.60&18.45\\
T0++-11b&Cal&4.24&3.34&9.48\\
flan-t5-xxl-11b&Eqa&3.74&5.78&43.89\\
flan-t5-xl-3b&\$&7.48&3.34&25.19\\
\hline
    \end{tabular}
    \caption{Evaluation on MATH 401 with different LLMs. Prompts are selected via best accuracy. Cal means ``Calculate:'' and Eqa means ``\textbackslash begin\{equation\}''. * means providing an additional system-level message.}
    \label{tab:main}
\end{table}

% \FloatBarrier
\begin{table}[t]
    \small
    \centering
    \begin{tabular}{lc|ccc}
    \hline
    Model & Prompt & Acc $\uparrow$ & RE $\downarrow$ & NNR $\downarrow$ \\
    \hline
% gpt-3.5-turbo-0301&\$\$&59.1&0.94&3.24\\
% gpt-3.5-turbo-0301&\$&62.34&0.91&3.24\\

% gpt-3.5-turbo-0301&Cal&\textbf{74.31}&\textbf{0.33}&\textbf{1.75}\\
% text-davinci-003&Cal&55.86&0.56&0.75\\
gpt-4&Cal*4&\textbf{83.54}&\textbf{0.07}&\textbf{0.00} \\
gpt-3.5-turbo-0301&Cal*&\underline{75.06}&\underline{0.14}&\underline{0.50}\\
text-davinci-003&Cal&\textit{56.61}&0.76&2.99\\
% text-davinci-002&\$\$&22.44&2.01&4.49\\
text-davinci-002&Cal&42.89&2.13&15.96\\
% text-curie-001&\$\$&8.23&2.20&1.50\\
% text-curie-001&Cal&11.72&1.73&1.75\\
text-curie-001&Cal&11.47&1.92&6.48\\
text-babbage-001&Eqa&5.24&2.59&5.74\\
code-davinci-002&Eqa&21.70&2.39&11.47\\
% text-davinci-002&Cal&43.64&1.3&3.74\\
% text-davinci-003&\$\$&43.39&1.11&0.75\\
% text-davinci-003&\$&45.64&1.19&1.0\\

% text-davinci-003&Equa&45.14&1.43&4.49\\
\hline
% Galactica-120b&\$\$&41.81&1.05&1.5\\
% Galactica-6.7b&\$\$&26.10&1.18&1.75\\
% galactica-120b&Eqa&43.72&0.91&1.0\\
% galactica-30b&Eqa&45.22&0.73&0.83\\
% galactica-6.7b&Cal&34.58&2.38&7.56\\

galactica-120b&Eqa&45.14&1.30&3.99\\
galactica-30b&Eqa&45.14&\textit{0.69}&\textit{1.75}\\
galactica-6.7b&Cal&34.41&2.61&8.73\\

%     \hline
% \textbf{AliGPT-175b-v24} & $\emptyset$ & 44.14&1.42&1.0\\
\hline
llama-65b&Eqa&28.43&1.61&4.74\\
llama-30b&Eqa&30.17&1.72&3.74\\
llama-13b&\$&27.68&2.40&9.73\\
llama-7b&\$\$&21.95&2.11&7.48\\
\hline
opt-175b&Cal&21.70&3.18&21.70\\
opt-66b&$\emptyset$&20.70&2.66&18.70\\
opt-iml-max-30b&Cal&17.46&1.52&6.23\\
opt-30b&$\emptyset$&15.96&2.28&11.22\\
opt-13b&$\emptyset$&15.21&2.19&10.97\\
opt-6.7b&Cal&14.46&1.46&4.24\\
% gpt-neox-20b&Eqa&35.83&0.79&1.58\\
% Gpt-j-6B&\$\$&24.19&1.35&4.57\\
% gpt-j-6b&Cal&28.26&1.66&8.89 \\
\hline
gpt-neox-20b&Eqa&35.41&1.19&4.49\\
gpt-j-6b&Cal&27.18&1.55&8.98\\
% \hline
% % \hline
% Alibaba-13B-0306&$\emptyset$&15.46&1.89&0.0\\
% Alibaba-13B-sftV16&$\emptyset$&16.96&3.09&1.25\\
\hline
% opt-iml-1.3b&Eqa&4.99&2.15&2.49\\
% opt-iml-30b&Eqa&17.46&1.25&0.25\\
% opt-iml-max-1.3b&Cal&4.24&1.38&0.75\\

% opt-66b&Cal&19.62&1.7&6.9\\
% opt-iml-max-30b&Cal&17.29&1.16&0.83\\
% opt-iml-30b&Eqa&16.87&1.16&0.25\\
% opt-30b&\$\$&14.96&2.98&12.97\\
% % opt-13b&\$\$&8.23&2.73&11.97\\
% opt-13b&Cal&13.47&1.68&3.66\\
% opt-6.7b&Cal&14.38&1.52&3.74\\
% opt-6.7b&\$\$&10.22&2.63&17.71\\
% opt-iml-max-1.3b&\$\$&3.24&2.62&13.97\\
% opt-iml-1.3b&\$\$&4.74&2.81&15.96\\

bloomz-176b&\$\$&22.44&1.50&4.74\\
bloom-176b&\$&20.2&2.60&18.45\\
bloomz-7b1&\$&12.72&2.56&15.46\\
bloom-7b1&Cal&7.23&2.41&6.48\\
bloomz-3b&\$\$&7.98&2.63&12.47\\
bloom-3b&Cal&4.24&2.41&8.73\\
bloomz-1b7&Eqa&4.74&4.28&31.17\\
bloom-1b7&Cal&5.24&2.54&11.22\\

% T0++&Cal&4.24&3.24&7.48\\
T0++-11b&Cal&4.24&3.34&9.48\\
\hline
glm-130b&\$&25.94&1.27&2.74\\
glm-10b&Cal&14.96&2.30&3.74\\
\hline
flan-t5-xxl-11b&Eqa&3.74&5.78&43.89\\
flan-t5-xl-3b&\$&7.48&3.34&25.19\\
flan-t5-large-780m&Cal&3.74&2.31&2.49\\
flan-t5-base-250m&Eqa&2.49&3.18&14.21\\

% flan-t5-base&Eqa&2.24&2.72&8.31\\
% flan-t5-large&Cal&3.49&2.38&2.74\\
% flan-t5-xl&\$&6.65&3.21&19.37\\
% flan-t5-xxl&Eqa&1.66&6.59&52.7\\
% flan-t5-xxl&\$\$&0.00&9.30&89.53\\
% flan-t5-xxl&Eqa&1.66&6.59&52.7\\
% % flan-t5-xl&\$\$&4.74&4.60&38.90\\
% flan-t5-xl&Cal&6.65&2.02&4.74\\
% % flan-t5-large&\$\$&2.24&5.04&41.4\\
% flan-t5-large&Cal&3.49&2.38&2.74\\
% flan-t5-base&Eqa&2.24&2.72&8.31\\
% google-flan-t5-base&Cal&1.75&2.0&4.49\\
\hline
    \end{tabular}
    \caption{Full evaluation on MATH 401 with different LLMs. Prompts are selected via best accuracy.}
    \label{tab:full}
\end{table}

\section{Results and Analysis}

\subsection{Results}
\paragraph{Overall Results} Table~\ref{tab:head},~\ref{tab:main}, and~\ref{tab:full} show results of different LLMs on MATH 401. We find GPT-4 and ChatGPT outperform all other models by a large margin\footnote{OpenAI states they improve the math of ChatGPT since version Jan 30, and we cannot evaluate any previous version.}.
GPT-4 surpasses ChatGPT with accuracy of 10 points and reduce relative error half.
InstructGPT performs third measured by accuracy and Galactica-30B performs third measured by relative error.
Compared to models proposed before InstructGPT (text-davinci-003), GPT-series applies Reinforcement Learning from Human Feedback (RLHF) which may enhance their arithmetic ability significantly.
Galactica is pre-trained with massive \LaTeX\space source codes which could be the reason why Galactica performs well in arithmetics.

\paragraph{Grouped Results} To clearly understand the arithmetic ability of LLMs, we show grouped accuracy in Table~\ref{tab:head}. 
GPT-4 obtains first places and ChatGPT obtains second places for all groups. 
Most LLMs are only capable of doing addition and subtraction and have some ability for multiplication.
Division, exponentiation, trigonometry functions, and logarithm functions are hard for most LLMs. LLMs have some abilities dealing with decimal, negative, and irrational numbers. Only GPT-4 and ChatGPT have the ability to deal with big numbers ($>1e12$) and complex long queries which proves their generalization and reasoning abilities. 
GPT-4 shows extremely good ability in long arithmetic expressions.
% GPT-4 xxx.
% ChatGPT achieves an accuracy of 99.5 in easy groups and 49 in hard groups which shows its powerfulness.

\paragraph{When will ChatGPT fail?} Though ChatGPT obtains such a good performance, we will check when ChatGPT fails to answer. For multiplication ($\times$), ChatGPT passes all queries in Group 7 and 8 and get wrong answers for all queries in Group 9. An example is ChatGPT predicts $71786\times21638=1,55\textcolor{red}{1},\textcolor{red}{4}0\textcolor{red}{2,0}68$, while the true answer is $1,553,305,468$. ChatGPT gives a very close estimation with the correct head and tail, which proves that ChatGPT does not use a calculator API for math calculation. 

For division in Group 11, ChatGPT sometimes gives correct answers to two decimal places which will be considered incorrect in our metric. We can see in Table~\ref{tab:system_prompt}, requiring ChatGPT to output four decimal places will improve its accuracy in multiplication and division.

For exponentiation ($\wedge$), ChatGPT correctly answers all queries in Group 10 which contain only integers as bases.
It is too hard for any language model (even ChatGPT) correctly estimate the exponentiation of a decimal number as the base and a decimal number as the exponent. It seems that ChatGPT treats $**$ as multiplication sometimes, for example, ChatGPT estimates $5.5507$ $**$ $2.0434=\textcolor{red}{10.31554}$ which is close to $5.5507\times2.0434=11.3423$ and far from answer $33.1895$.

For calculating trigonometry functions, ChatGPT understands degrees and radians correctly and generates exact answers for special inputs like $cos(-210^{\circ})=-\frac{\sqrt{3}}{2}$ (we omit explanation generated by ChatGPT here). However, ChatGPT may generate wrong explanations which mislead itself. An example is: ``We know that the sine function is periodic with a period of $2\pi$, which means that $\sin(x + 2\pi) = \sin(x)$ for any value of $x$. Therefore, we can subtract multiples of $2\pi$ from $-3.75\pi$ until we get a value between 0 and $2\pi$: $-3.75\pi = -3\pi - 0.75\pi = -9.42477 - 2.35619 = -11.78096$. Adding 2$\pi$, we get: $-11.78096 + 2\pi$ = \textcolor{red}{-9.42477} etc.'' Any mistake in explanations may result in a wrong answer.

For logarithm functions, we find that ChatGPT is capable of using change of base formula and predicting answers within two decimal places.

For long expressions, ChatGPT can understand the operators' priorities. ChatGPT sometimes generates answers step by step and sometimes generates answers directly.
It is very likely to generate wrong answers when it decodes answers directly.

\paragraph{What about GPT-4?}

For big number multiplication ($\times$) in group 9, GPT-4 also fails in all cases with similar problems occurring in ChatGPT.

For exponentiation ($\wedge$), GPT-4 will not consider $**$ as $\times$ anymore and give better estimations.

For calculating expressions with irrational numbers, GPT-4 will consider $e$ as natural logarithm correctly.

For logarithm functions, GPT-4 calculates logarithm base $e$ and 10 by ``using a calculator'' (this is a message generated by GPT-4). GPT-4 calculates logarithm base 2 by change of base formula and generates approximate results.

For long equations, GPT-4 solves all equations step by step and obtains a much higher accuracy.

We compare and summarize how GPT-4 outperforms ChatGPT here:
\begin{itemize}
    \item Better division ability.
    \item Better trigonometry ability.
    \item Understand irrational numbers properly.
    \item Always calculate long expressions step by step.
\end{itemize}

% \begin{table*}[h]
%     \centering
%     \begin{tabular}{ll|cccc}
%     Best Prompt & Model & Accu.(\%)  & AVG. ERR & NON NUM. RATE \\
%     \hline
%     Calculate:&google/flan-t5-xxl&0.0&0.0&10.0&1.0\\
%     Calculate:&google/flan-t5-xl&0.0&2.99&358.18&0.09\\
%     Calculate:&google/flan-t5-large&0.0&5.8&318.7&0.0\\
%     Calculate:&google/flan-t5-base&0.0&3.77&408.48&0.06\\
%     Calculate:&google/flan-t5-small&0.0&3.89&518.32&0.02\\
%     Calculate:&UL2\\

%     Calculate:&bigscience/bloom-560m&0.0&1.7&23.7&0.01\\
%     Calculate:&bigscience/bloom-1b1&0.0&0.75&116.24&0.0\\

%     Calculate:&bigscience/bloomz-1b1&0.0&4.26&9546.96&0.01\\
%     Calculate:&bigscience/bloomz-3b&3.0&5.06&9568.96&0.02\\
%     Calculate:&facebook/opt-iml-max-13b\\
%     Calculate:&facebook/opt-13b&2.0&1.35&104.54&0.0\\
%     Calculate:&facebook/opt-6.7b&3.0&2.08&193.33&0.0\\
%     Calculate:&facebook/opt-2.7b&0.0&1.71&124.84&0.0\\
%     Calculate:&facebook/opt-1.3b&0.0&1.85&98.34&0.0\\
%     Calculate:&facebook/opt-iml-max-1.3b\\
%     Calculate:&GPT-J\\
%     Calculate:&GPT-NeoX\\
%     Calculate:&text-davinci-003\\
%     Calculate:&gpt-3.5-turbo-0301\\
%     Calculate:&code-davinci-002\\
%     Calculate:&GLM\\
%     Calculate:&LamDa\\
%     Calculate:&Galactica\\
%     Calculate:&T0++\\
%     Calculate:&J1-Jumbo v1 (178B)\\
    
%     \end{tabular}
%     \caption{Caption}
%     \label{tab:my_label}
% \end{table*}

\subsection{Tokenization}
\label{sec:Tokenizer}
Arithmetic expressions have special tokens including $\pi, \times, \div, {}^{\circ}$ which are not within T5 series models (i.e. T0++ and Flan-T5).
T0++-11B (Acc 4.24 and RE 3.34) and Flan-T5-xxl-11B (Acc 3.74 and RE 5.78) perform badly on arithmetic tasks compared to other similar-size models: Opt-13B (Acc 15.21 and RE 2.19) and LLaMA-13B (Acc 27.68 and RE 2.4).

We notice that Galactica and LLaMA split numbers into individual tokens. For example $123.456$ is converted into $1\ 2\ 3\ .\ 4\ 5\ 6$. \citet{Razeghi2022ImpactOP} show that arithmetic ability is related to pre-training term frequencies. For tokens that appear more in pre-training, LLMs can have better accuracy in answering arithmetic expressions about them.
Number tokens with more digits (e.g. 23) apparently appear less than single digit token (e.g. 2 and 3).
Splitting numbers into individual tokens neglects all number tokens with more digits and makes all single digit tokens (mainly $0\sim 9$) appear in the pre-training corpus in the same order of magnitude.
Galactica-30B and LLaMA-30B obtain 45.14 and 30.17 in terms of accuracy (list in Table~\ref{tab:full}) that outperforms OPT-30B (15.96), Bloom-176B (20.2), and GLM-130B (25.94), which show superiority of digit-level tokenization.

% Continuous tokens for numbers 

\subsection{Training}
\label{sec:pretrain}
\paragraph{Self-supervised Pre-training} While pre-training, code corpus and \LaTeX-sources are possible to relate to arithmetic ability since they all contain arithmetic operators and numbers. Code-davinci-002 is pretrained with code corpus. Code-davinci-002 performs well on many reasoning-related tasks \cite{Zhou2022LeasttoMostPE}, however, it performs not good compared to other LLMs in arithmetics. This proves that mathematical reasoning ability is different from arithmetic ability which needs to understand numbers deeply.
Galactica with numerous \LaTeX-sources outperforms other LLMs except for InstructGPT and ChatGPT which show \LaTeX\space is useful.

\paragraph{Instruction Tuning} is also very important in arithmetic ability. Comparing Opt-30B (Acc 15.96 RE 2.28 NNR 11.22) with Opt-Iml-Max-30B (Acc 17.46 RE 1.52 NNR 6.23), Bloom (Acc 20.2 RE 2.6 NNR 18.45) with BloomZ (Acc 22.44 RE 1.5 NNR 4.74), and code-davinci-002 (Acc 21.7) with text-davinci-002 (Acc 42.89) in Table~\ref{tab:full} show that instruction tuning can boost the performance in all metrics.
Text-davinci-003 (RLHF) outperforms text-davinci-002 (SFT) in arithmetic tasks which shows RLHF is important for building arithmetic ability.

\subsection{Prompts}
\label{sec:prompt}

\paragraph{Input Prompts}
We find the best prompts are different across LLMs. We list the best and worst prompts for LLMs in Table~\ref{tab:prompt}.
We find models are sensitive to input prompts and not using prompts is the worst option for most LLMs. For InstructGPT and ChatGPT, using ``Calculate'' as a prompt perform best. For other LLMs, using \LaTeX-related prompts perform best.

\begin{table}[t]
    \small
    \centering
    \begin{tabular}{l|cccc}
  \hline
    Model & Best & Acc & Worst & Acc \\
  \hline
gpt-3.5-turbo-0301&Cal*&\textbf{75.06}&\$\$&64.59\\
text-davinci-003&Cal&\underline{56.61}&Eqa&43.64\\
galactica-120b&Eqa&45.14&$\emptyset$&38.9\\
llama-65b&Eqa&28.43&Cal&4.74\\
opt-175b&Cal&21.7&$\emptyset$&15.21\\
gpt-neox-20b&Eqa&35.41&$\emptyset$&26.93\\
glm-130b&\$&25.94&$\emptyset$&22.44\\
bloomz-176b&\$\$&22.44&$\emptyset$&11.72\\
  \hline
    \end{tabular}
    \caption{Best and worst prompts for different LLMs.}
    \label{tab:prompt}
\end{table}

\paragraph{System Prompts}
For ChatGPT, we can also provide system-level messages as instruction prompts. 
Table~\ref{tab:system_prompt} shows providing system-level messages improves ChatGPT's accuracy and reduces relative error significantly.
The most different groups are group 13 irrational numbers and group 16 logarithm functions. Without a system-level message, ChatGPT thinks $e$ can be Euler's number or a variable and cannot give an answer.
For logarithm functions, ChatGPT tries to explain how it calculates which may mislead our provided parser.
We notice that if we require ChatGPT to output results to four decimal places, it will have a zero non-number ratio.
To conclude, ChatGPT will try to explain the calculation procedure without a system-level prompt and will only provide answers with a system-level prompt.

\begin{table}[h]
    \small
    \centering
    \begin{tabular}{l|cccccc}
  \hline
  Group  & \multicolumn{2}{c}{Cal} & \multicolumn{2}{c}{Cal*} & \multicolumn{2}{c}{Cal*4} \\
  &Acc&RE&Acc&RE&Acc&RE\\
  \hline
   0 Euler & \textbf{100} & \textbf{.00}&\textbf{100} & \textbf{.00}&\textbf{100} & \textbf{.00}\\
   $1\sim6$ $+-$ & \textbf{97} & \textbf{.00} & 96 & .00 & 93 & .01 \\
   $7\sim10$ $\times\div$ &69&.20&69&\textbf{.01}&\textbf{71}&\textbf{.01} \\
   $11\sim12$ $\wedge$ &\textbf{50}&\textbf{.24}&\textbf{50}&.32&\textbf{50}&.27 \\
   $13$ Irr.&64&1.73&72&.56&\textbf{84}&\textbf{.11} \\
   $14$ Long&\textbf{68}&\textbf{.19}&64&.46&60&.59 \\
   $15$ Tri.&44&1.21&\textbf{48}&\textbf{.96}&44&1.40 \\
   $16$ Log&56&.80&\textbf{60}&.04 &56&\textbf{.01}\\
  \hline
    Overall & 74&.33&\textbf{75}&\textbf{.14}&74&\textbf{.14}\\
   \hline
    \end{tabular}
    \caption{Comparing different system prompts in ChatGPT on MATH 401. Cal means no system prompt. * = ``You are an accurate calculator.'' 4 = ``Calculating to four decimal places.''}
    \label{tab:system_prompt}
    \centering
\end{table}

% \paragraph{Query Rewriting} We notice that GPT-4 is re-writing expressions

% Estimate

% Estimate it step by step

% Estimate it using its expansion

% \subsection{Prompt Probing}
% When pre-training, a common trick is to combine several input sequences into one long sequence to fulfill the GPU memory usage.
% When we require models to answer an arithmetic expression, models generate answers first. Models may generate another instruction and arithmetic expression related. These instructions may be used in their models' multi-task instruction fine-tuning. We list their possible instructions used for instruction fine-tuning in Table~\ref{tab:probe}.

% \begin{table}[h]
%     \small
%     \centering
%     \begin{tabular}{l|p{5.5cm}}
%   \hline
%   Model & Prompt \\
% \hline 
%   % Flan \\
%   Bloom & You can use the following code to get the sum of the numbers in:\\
%   OPT-IML & What is the value of ...\\
% \hline
%     \end{tabular}
%     \caption{Detected possible prompts used for LLMs.}
%     \label{tab:probe}
%     \centering
% \end{table}

% % \subsection{Decode Strategy}

% % Arithmetic tasks are different from other natural language tasks which have multiple acceptable results.

% \FloatBarrier
\begin{table*}[t]
    \small
    \centering
    \begin{tabular}{p{2cm}|p{6.5cm}p{6.5cm}}
    \hline
    Expression & Calculate: & Zero-shot COT \\
    (88 - 42) * (33 * 41) - (13 - 27) = 62252&
    First, we need to simplify the expression inside the parentheses: 88 - 42 = 46 33 * 41 = 1353 13 - 27 = -14 Now we can substitute these values into the original expression: (46) * (1353) - (-14) Multiplying 46 and 1353 gives: 62,238 Remember that subtracting a negative number is the same as adding a positive number, so we can rewrite the expression as: 62,238 + 14 Adding 14 gives: 62,252 Therefore, the answer is: 62,252 & First, we simplify the parentheses: (88 - 42) * (33 * 41) - (13 - 27) = 46 * 1353 - (-14) Next, we simplify the second set of parentheses by changing the subtraction into addition of the opposite: = 46 * 1353 + 14 We can then multiply 46 and 1353: = \textcolor{red}{62338} + 14 Finally, we add: = \textcolor{red}{62352} Therefore, (88 - 42) * (33 * 41) - (13 - 27) = \textcolor{red}{62352}. \\
    \hline
    \end{tabular}
    \caption{Example of Group 14 decoded by ChatGPT with Calculate and COT prompts.}
    \label{tab:cot_example}
    \centering
\end{table*}

\subsection{Interpolation and Extrapolation}
\label{sec:inter}
LLMs have strong abilities to fit on in-domain data.
If pretraining corpora contain arithmetic expressions, it is easy for LLMs to memorize them.
For out-of-domain data, LLMs need to extrapolate how to calculate them.
% and are easy to perform well on simple arithmetic tasks like single-digit addition.
% If the pretraining corpora contain the exactly same arithmetic expressions, it is possible that LLMs decode the correct answers by memorizing. 
We do not know what are in-domain data and out-of-domain data for models (especially ChatGPT), so it is hard to test their interpolation and extrapolation abilities.
% the exact instruct-tuning data used in InstructGPT and ChatGPT, so we cannot say which part of MATH401 can 
We use the easy group and the hard group to estimate the interpolation and extrapolation abilities.
% If the pretraining corpora contain 
The easy group queries have possibilities that appear in the pretraining corpora or instruct-tuning, while the hard group queries contain big numbers / decimal numbers / long expressions which are very unlikely to be covered by pretraining corpora or instructions. Thus answering easy queries may examine the interpolation ability of models and answering hard queries must examine the extrapolation ability of the models.
We find ChatGPT performs best on hard queries, and all other models have limited performance on hard queries which show limited extrapolation.

\begin{figure}[t]
    \centering
    \includegraphics[width=0.53\textwidth]{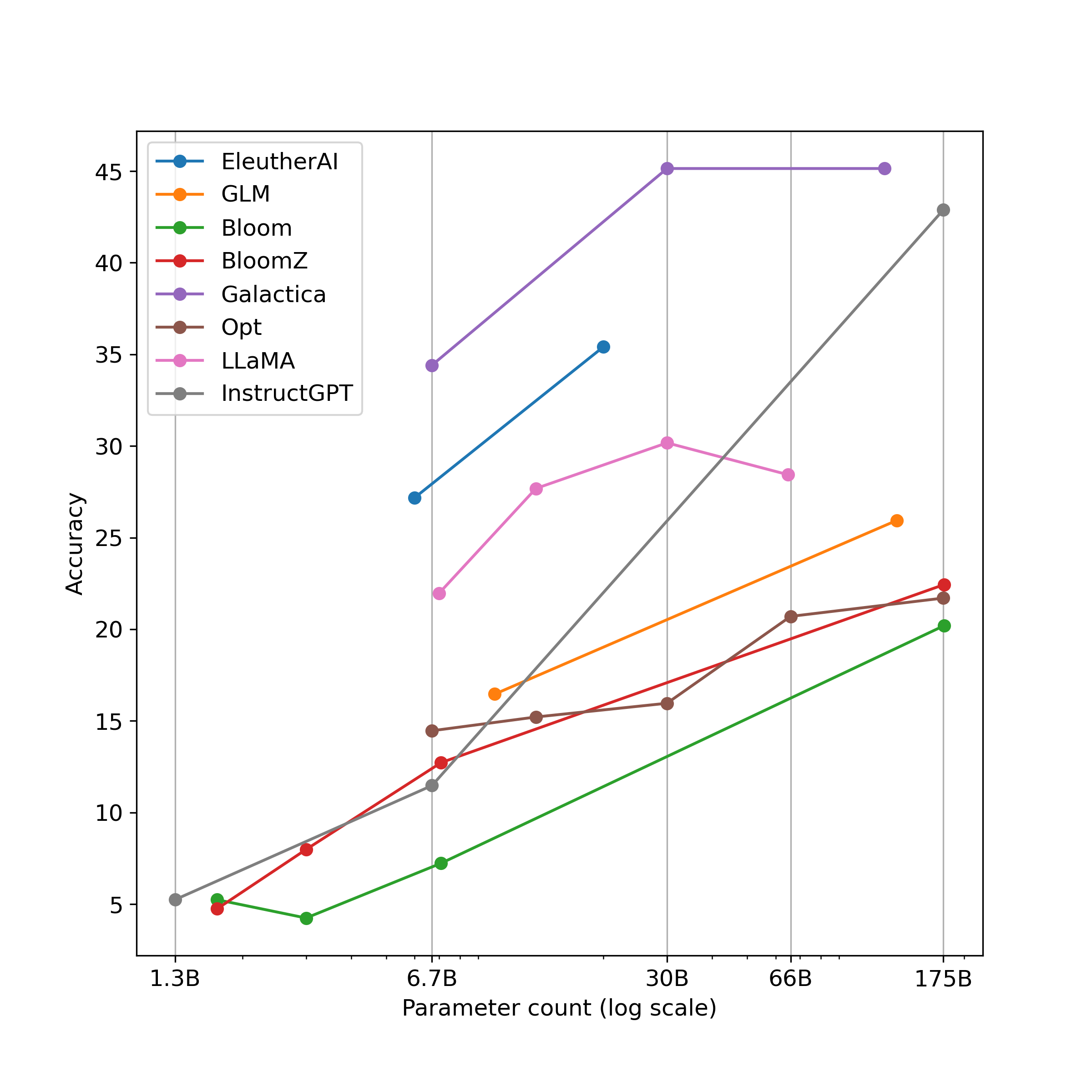}
    \caption{Performances of MATH 401 on LLMs with different sizes. We do not know the parameter count of ChatGPT. We list InstructGPT results with SFT setting (text-davinci-002) only for a fair comparison.}
    \label{fig:sc}
\end{figure}

\subsection{Scaling Laws}
\label{sec:sc}

To understand how parameter counts influence arithmetic ability, we plot the results with different-size LLMs in Figure~\ref{fig:sc}.
We do not plot text-davinci-003, gpt-3.5-turbo-0301 and gpt-4 since they do not have smaller versions with the same setting.
We find that LLMs have better abilities with larger parameter counts.
An interesting phenomenon we found is model over 30B does not improve significantly compared with 30B models, especially in Galactica where the 120B model performs the same as the 30B model. We hypothesize that 30B may be enough for arithmetic ability. 
% There are some guesses that ChatGPT is less than 175B which 
ChatGPT may be a model smaller than 175B which outperforms other 175B models a lot, thus larger parameter count does not guarantee better arithmetic ability.
For GPT-4, we cannot have any possible guess. Considering its much slower inference speed, we guess it has larger parameter counts than ChatGPT and obtain better reasoning ability (i.e. long arithmetic expression).
% Another piece of evidence is ChatGPT may be smaller than 175B which can 

\subsection{Chain-of-Thought}
\label{sec:cot}

% \paragraph{Zero-shot COT}
LLMs can leverage chain-of-thought to better answer math word problems \cite{cot}.
We test on ChatGPT whether chain-of-thought will improve arithmetic calculations.
We use the prompt ``Let us solve this equation step by step'' to instruct ChatGPT for zero-shot COT \cite{0shotcot}.
We compare the results of zero-shot COT using ``Calculate:'' in Table~\ref{tab:cot}.
Surprisingly, we find that COT does not improve the performance of any group even in group 14 with long arithmetic expressions. To understand the reason for this phenomenon, we check decoded results for these two prompts in Table~\ref{tab:cot_example}.
We find using ``Calculate:'' as the prompt can automatically generate chain-of-thoughts for long arithmetic expressions and generate answers directly for easy questions.
% (1) Use calculate also generate zero-shot COT; (2) Use COT prompt, the model tends to output a fraction instead of a decimal number.

\begin{table}[h]
    \small
    \centering
    \begin{tabular}{l|cccc}
  \hline
  Group  & \multicolumn{2}{c}{Cal} & \multicolumn{2}{c}{0 COT} \\
  &Acc&RE&Acc&RE\\
  \hline
   0 Euler & \textbf{100} & \textbf{.00}&\textbf{100} & \textbf{.00}\\
   $1\sim6$ $+-$ & \textbf{97} & \textbf{.00} & 94 & .02 \\
   $7\sim10$ $\times\div$ &\textbf{69}&\textbf{.20}&61&.66 \\
   $11\sim12$ $\wedge$ &\textbf{50}&\textbf{.24}&48&.56 \\
   $13$ Irr.&\textbf{64}&\textbf{1.73}&28&4.89 \\
   $14$ Long&\textbf{68}&\textbf{.19}&64&.46 \\
   $15$ Tri.&\textbf{44}&1.21&40&\textbf{1.14} \\
   $16$ Log&\textbf{56}&\textbf{.80}&28&5.37 \\
  \hline
    Overall & \textbf{74} & \textbf{.33} & 66 & .98\\
\hline 
    \end{tabular}
    \caption{Comparing zero-shot COT and Calculate using ChatGPT on MATH 401.}
    \label{tab:cot}
    \centering
\end{table}

\begin{table}[t]
    \small
    \centering
    \begin{tabular}{l|cccc}
  \hline
    Model & \multicolumn{2}{c}{Naive} & \multicolumn{2}{c}{ICL} \\
    &Acc&RE&Acc&RE\\
  \hline
galactica-120b&45.14&1.3&45.14&0.42\\
galactica-6.7b&34.41&2.61&32.67&0.65\\
flan-t5-xxl&3.74&5.78&0.0&10.0\\
flan-t5-base&2.49&3.18&0.0&10.0\\
  \hline
    \end{tabular}
    \caption{In-context learning on MATH 401.}
    \label{tab:prompt}
\end{table}

\subsection{In-context Learning}
\label{sec:icl}
In-context learning (ICL) provides related question-answer pairs to improve LLMs \cite{gpt3,cot}. In our task, we can provide similar arithmetic expressions before the queries to help model understanding the arithmetic operator as done in \citet{Smith2022UsingDA}. We provide 8 similar cases (we promise these cases are different from the query) for each query.
We test whether ICL can improve the well-behaved model (Galactica) and the underperforming model (Flan-T5).
For Galactica, it does not improve accuracy but reduces relative error significantly.
For small-sized Flan (smaller than 3B) it cannot generate any number under the setting of in-context-learning.

\section{Conclusion}
In this paper, we propose MATH 401 to evaluate the arithmetic ability of LLMs. We find that tokenization, pre-training corpus, prompts, and model parameter counts are important for their arithmetic ability. The reason ChatGPT performs so well in arithmetic still has some mystery, i.e. the parameter counts and instruction datasets of ChatGPT. We hope this paper can help readers improve LLMs with better arithmetic ability.
This paper is only focused on arithmetic, testing LLMs on other math topics including symbolic mathematics, solving (ordinary differential, partial differential) equations, calculus, algebra, geometry, probability theory, and graph theory are also interesting topics.

\bibliography{custom,anthology}
\bibliographystyle{acl_natbib}

\appendix

\section{Examples from MATH 401}
\label{sec:app:math401}
We list examples for each group from MATH 401.

\begin{itemize}
    \item $e^{i\pi}+1=0$
    \item $5+9=14$
    \item $21+97=118$
    \item $721-847=-126$
    \item $714637232158-667119914538=47517317620$
    \item $-1+(-6)=-7$
    \item $-0.038+0.0092=-0.0288$
    \item $78 \times 64=4992$
    \item $5.0 \times 0.09=0.045$
    \item $45960 \times 59693=2743490280$
    \item $70 \div 61=1.1475$
    \item $7^4=2401$
    \item $2.242^{3.7342}=20.3865$
    \item $e+\pi=5.8598$
    \item $( 4  \times  64)  \times  ( 39 + 12)=13056$
    \item $\sin(-3.75\pi)=0.7071$
    \item $\log_{10}(797)=2.9015$
\end{itemize}

% \section{Evaluation Results for LLMs in all sizes}

% \section{Examples of Group 14 decoded by ChatGPT}

\end{document}